\documentclass{article}
\usepackage{spconf,amsmath,graphicx,comment}
\usepackage{algorithm}
\usepackage[pdf]{pstricks}
\usepackage{auto-pst-pdf}
\usepackage{psfrag}
\usepackage{algpseudocode,textcomp}
\algtext*{EndWhile}
\algtext*{EndFor}

\title{A HYBRID RECURRENT NEURAL NETWORK FOR MUSIC TRANSCRIPTION}
\nametwo{Siddharth Sigtia$^\ast$\thanks{SS is supported by a City University London Pump-Priming Grant; EB is supported by a City University London Research Followship;NB is currently working at Google Inc, Mountain View, California, USA},Emmanouil Benetos$^\dag$,Nicolas Boulanger-Lewandowski$^\ddag$,Tillman Weyde$^\dag$}{Artur S. d'Avila Garcez$^\dag$ and Simon Dixon$^\ast$}
\address{$\ast$Centre for Digital Music, EECS, Queen Mary University of London, London, UK\\
$\dag$ Department of Computer Science, City University London, London, UK\\
$\ddag$ Dept. IRO, Universit\'{e} de Montr\'{e}al, Montr\'{e}al (QC), H3C 3J7, Canada}
\begin{document}
\ninept
\maketitle
\begin{abstract}
We investigate the problem of incorporating higher-level symbolic score-like information into Automatic Music Transcription (AMT) systems to improve their performance. We use recurrent neural networks (RNNs) and their variants as music language models (MLMs) and present a generative architecture for combining these models with predictions from a frame level acoustic classifier. We also compare different neural network architectures for acoustic modeling. The proposed model computes a distribution over possible output sequences given the acoustic input signal and we present an algorithm for performing a global search for good candidate transcriptions. The performance of the proposed model is evaluated on piano music from the MAPS dataset and we observe that the proposed model consistently outperforms existing transcription methods. 
\end{abstract}
\section{Introduction}
\label{sec:intro} 

Automatic Music Transcription (AMT) involves identifying the pitches present in a given polyphonic acoustic signal and generating a corresponding symbolic, score-like transcription \cite{Klapuri06book}. Most AMT systems focus primarily on modeling the acoustic signal to identify the pitches present as a function of time. Music exhibits structural regularity much like language, and therefore symbolic music prediction systems or Music Language Models (MLMs) can provide accurate symbolic priors and have the potential to significantly improve AMT systems. However, MLMs have not been extensively applied to AMT because polyphonic symbolic music prediction is quite a difficult problem and simple models such as n-grams which are used in speech are insufficient for modeling sequences of polyphonic music \cite{boulanger2012modeling}. 

Recurrent neural networks (RNNs) are powerful temporal models that can, in theory, capture long-term dependencies between inputs because of their powerful hidden representation. RNNs and their more complex variants \cite{boulanger2012modeling}, have recently been applied successfully to the problem of symbolic music prediction. This has led to a revival of interest in the problem of incorporating prior symbolic knowledge to improve AMT systems. Although RNNs achieve reasonable accuracy at symbolic music prediction tasks, it is not obvious how these priors can be incorporated into music transcription systems. The obvious strategy of multiplying the predictions of the acoustic and language models and then renormalizing, like in a product of experts, suffers from the \emph{label bias} problem for low entropy sequences \cite{Lafferty:2001:CRF:645530.655813}. 

Recently, there have been a few studies that try to incorporate symbolic priors into AMT systems. The model proposed in \cite{boulanger2013high}, is an input-output variant of the RNN-RBM model for music transcription. Although the model performs well on several datasets, it suffers from the problem of \emph{teacher forcing}, where the acoustic and symbolic information are incorrectly weighted. The system in \cite{raczynski2013dynamic}, uses a family of Dynamic Bayesian Network (DBN) language models to complement the acoustic model, though the search space of possible transcriptions must be constrained in order for the method to be tractable. In \cite{simsekli2013hierarchical}, the authors propose a novel dynamical system for incorporating symbolic information into a non-negative factorisation based transcription model. The method proposed in \cite{sigtiarnn}, incorporates symbolic information into a PLCA based transcription system using Dirichlet priors. Although the model performs well, it can only be used when the acoustic model is based on spectrogram factorisation techniques. Another shortcoming of the model in \cite{sigtiarnn} is that the acoustic and language models are trained independently by optimising different objectives. 

The popular technique of superposing a Hidden Markov Model (HMM) to the outputs of a frame-level classifier, like in state-of-the-art speech recognition systems \cite{hinton2012deep} is intractable for AMT tasks. This is because the outputs of the acoustic classifier at any time are high-dimensional binary vectors. Consequently, the number of hidden HMM states is exponential in the number of output variables. This makes the parameter estimation problem for the HMM intractable. HMMs can be applied to polyphonic AMT systems under the assumption that each pitch is independent of all the other pitches \cite{poliner2006discriminative}. However this assumption is violated by polyphonic music and therefore the method is unsatisfactory. 

In this paper we employ the architecture in \cite{boulangerphone}, which was originally proposed for modelling sequences of phonemes in speech recognition. The architecture provides a principled way for superposing an RNN to the predictions of an \emph{arbitrary} frame level classifier and combines the two models under a common training objective. It is advantageous to use RNNs for high-dimensional problems like AMT, since the outputs of the RNN form a distributed representation, which makes the parameter estimation problem more tractable compared to an HMM. Additionally, the predictions of an RNN are conditioned on the entire sequence history which is a generalisation over the HMM transitions which are conditioned only on the previous time-step. We also compare performance between using Deep Neural Network (DNN) and RNN acoustic models. We present an efficient high-dimensional beam-search algorithm for decoding and compare the performance of this \emph{hybrid} architecture to existing AMT systems.

The rest of the paper is organised as follows. Section 2 introduces RNNs. Section 3 describes the hybrid architecture and Section 4 discusses the inference algorithm that is used for testing. Section 5 describes the experimental setup and details of training. Section 6 discusses the results and the paper is concluded in Section 7. 

\section{Recurrent Neural Networks}
\label{sec:format}
An RNN is a powerful discrete-time dynamical system that can in principle, capture complex long term dependencies between its inputs. An RNN, when used as a generative model, defines a distribution over a sequence $z$ in the following manner:
\begin{equation}
P(z) = \prod_{t=1}^{T} P(z_t|\mathcal A_t)
\end{equation} where $\mathcal A_t \equiv \left\{ z_{\tau} | \tau < t \right\}$ is the sequence history at time $t$. The hidden state of an RNN with a single layer of hidden units is defined by the following recurrence relation:
\begin{equation}
h_t = \sigma(W_{zh}z_{t-1} + W_{hh}h_{t-1} + b_{h})
\end{equation} where $W_{zh}$ are the weights from the inputs at $t-1$ to the hidden units at $t$, $W_{hh}$ are the recurrent weights between hidden units at $t-1$ and $t$ and $b_h$ are the hidden biases. 

The output vector at time $z_t$ is obtained in the following way:
\begin{equation}
 z_t = f(W_{hz}h_t + b_z)
\end{equation} where $f$ is some function applied to each element. The choice of $f$ depends on the outputs that are being modeled. If the output variables form a one-of-K representation, then $f$ is a softmax function that yields a multinomial distribution at the outputs. When $f$ is a sigmoid function, then the outputs represent the independent probabilities of occurrence of each output variable. 

The fact that the output variables are independent of each other is a very restrictive assumption when used for modeling polyphonic music. This is because musical notes appear in highly correlated patterns where the presence or absence of a note influences the likelihood of occurrence of all other notes. Therefore, instead of using the RNN to predict the probabilities of pitches directly, we can use the RNN to predict the parameters of a high-dimensional distribution estimator like the Restricted Boltzmann Machine (RBM) or the Neural Autoregressive Density Estimator (NADE) \cite{boulanger2012modeling}. The RNN-NADE is a natural choice for a language model since it is tractable to obtain probabilities from the conditional NADEs at each step, which is necessary during inference. Another advantage of using the RNN-NADE is that the gradients of the objective function can be calculated exactly and therefore we can make use of more powerful optimisers like Hessian Free (HF) \cite{Martens2011}.

\section{Hybrid Architecture}

In this section we describe the architecture used to combine an RNN-based MLM with an arbitrary frame level classifier. The architecture is a generative graphical model that generalises the HMM architecture by conditioning predictions at some time $t$, on all previous predictions $\tau < t$, as opposed to the HMM, where $\tau = t-1$. Figure 1 is a graphical representation of the architecture.

The hybrid architecture factorises the joint probability of the sequence of acoustic vectors $x$ and their corresponding labels $z$ in the following way:

\begin{align}
 P(z,x) &= P(z_{1} \ldots z_{T} , x_{1} \ldots x_{T}) \\ 
  &= P(z_1)P(x_{1}|z_1) \prod_{t=2}^{T} P(z_t|\mathcal{A}_t) P(x_t|z_t). \\ \nonumber
\end{align}

In the above factorisation, the symbolic prediction terms $P(z_t|\mathcal{A}_t)$ can be obtained from an RNN, while the $P(x_t|z_t)$ terms are \emph{emission} probabilities of observing the acoustic vector $x_t$ given a state $z_t$. The above factorisation makes the following independence assumption for an emitted acoustic vector $x_t$:
\begin{equation}
\label{independence}
P(x_t | z, \left\{ x_{\tau} , \tau < t \right\}) = P(x_t | z_t). 
\end{equation}
Using Bayes' rule, the joint probability can be reformulated in terms of the scaled likelihood:

\begin{equation}
P(z,x) \propto P(z_1;\Theta_l) \frac{P(z_1|x_{1})}{P(z_1)} \prod_{t=2}^{T} P(z_t|\mathcal{A}_t) \frac{P(z_t|x_t)}{P(z_t)}
\end{equation} where $\Theta_l$ are the parameters of the language model. 
The term $P(z_t|x_t)$ can be obtained from the output of an arbitrary frame-level classifier, $P(z_t)$ is the marginal distribution of target vectors which can be easily calculated from the training set and constant terms involving $x_t$ have been removed by introducing the proportionality symbol. 

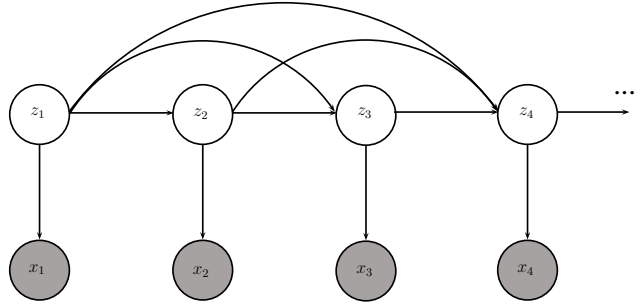
\begin{figure}[]
\begin{minipage}[b]{1.0\linewidth}
  \centering
  \resizebox{240pt}{!}{
\scalebox{1} 
{
\begin{pspicture}(0,-0.8)(15.07875,6.44)
\definecolor{color479b}{rgb}{0.6549019607843137,0.6352941176470588,0.6352941176470588}
\pscircle[linewidth=0.04,dimen=outer](0.74,3.72){0.74}
\pscircle[linewidth=0.04,dimen=outer,fillstyle=solid,fillcolor=color479b](0.74,-0.04){0.74}
\psline[linewidth=0.04cm,arrowsize=0.05291667cm 2.0,arrowlength=1.4,arrowinset=0.4]{->}(0.74,3.0)(0.74,0.7)
\psline[linewidth=0.04cm,arrowsize=0.05291667cm 2.0,arrowlength=1.4,arrowinset=0.4]{->}(1.46,3.76)(3.98,3.76)
\pscircle[linewidth=0.04,dimen=outer](4.68,3.72){0.74}
\pscircle[linewidth=0.04,dimen=outer,fillstyle=solid,fillcolor=color479b](4.68,-0.04){0.74}
\psline[linewidth=0.04cm,arrowsize=0.05291667cm 2.0,arrowlength=1.4,arrowinset=0.4]{->}(4.68,3.0)(4.68,0.7)
\psline[linewidth=0.04cm,arrowsize=0.05291667cm 2.0,arrowlength=1.4,arrowinset=0.4]{->}(5.4,3.76)(7.92,3.76)
\pscircle[linewidth=0.04,dimen=outer](8.62,3.7){0.74}
\pscircle[linewidth=0.04,dimen=outer,fillstyle=solid,fillcolor=color479b](8.62,-0.06){0.74}
\psline[linewidth=0.04cm,arrowsize=0.05291667cm 2.0,arrowlength=1.4,arrowinset=0.4]{->}(8.62,2.98)(8.62,0.68)
\usefont{T1}{ptm}{m}{n}
\rput(0.704375,3.765){\large $z_1$}
\usefont{T1}{ptm}{m}{n}
\rput(4.6389065,3.745){\large $z_2$}
\usefont{T1}{ptm}{m}{n}
\rput(8.591406,3.765){\large $z_3$}
\usefont{T1}{ptm}{m}{n}
\rput(0.685625,-0.015){\large $x_1$}
\usefont{T1}{ptm}{m}{n}
\rput(4.6401563,-0.075){\large $x_2$}
\usefont{T1}{ptm}{m}{n}
\rput(8.592656,-0.055){\large $x_3$}
\usefont{T1}{ptm}{m}{n}
\rput(14.86375,4.25){\huge ...}
\psline[linewidth=0.04cm,arrowsize=0.05291667cm 2.0,arrowlength=1.4,arrowinset=0.4]{->}(9.3,3.78)(11.82,3.78)
\pscircle[linewidth=0.04,dimen=outer](12.52,3.72){0.74}
\pscircle[linewidth=0.04,dimen=outer,fillstyle=solid,fillcolor=color479b](12.52,-0.04){0.74}
\psline[linewidth=0.04cm,arrowsize=0.05291667cm 2.0,arrowlength=1.4,arrowinset=0.4]{->}(12.52,3.0)(12.52,0.7)
\psline[linewidth=0.04cm,arrowsize=0.05291667cm 2.0,arrowlength=1.4,arrowinset=0.4]{->}(13.24,3.78)(14.98,3.78)
\usefont{T1}{ptm}{m}{n}
\rput(12.500156,3.765){\large $z_4$}
\usefont{T1}{ptm}{m}{n}
\rput(12.481406,-0.035){\large $x_4$}
\psarc[linewidth=0.04,arrowsize=0.05291667cm 2.0,arrowlength=1.4,arrowinset=0.4]{<-}(4.68,1.68){3.82}{33.448994}{147.1491}
\psarc[linewidth=0.04,arrowsize=0.05291667cm 2.0,arrowlength=1.4,arrowinset=0.4]{<-}(8.6,1.7){3.82}{33.448994}{147.1491}
\psarc[linewidth=0.04,arrowsize=0.05291667cm 2.0,arrowlength=1.4,arrowinset=0.4]{<-}(6.64,0.0){6.42}{36.14204}{143.95937}
\end{pspicture} 
}}
\end{minipage}
\label{fig:res}
\caption{Proposed hybrid architecture.}
\end{figure}

We train the model by maximising the log-likelihood of occurrence of pairs of training examples $x,z$. The model can be easily trained with gradient descent because the gradient of the log-likelihood splits up into terms associated with the acoustic and language models in the following way:
\begin{equation}
\label{train_1}
\frac{\partial \log P(x,z)}{\partial \Theta_a} = \frac{\partial}{\partial \Theta_a} \sum_{t=1}^T \log P(z_t|x_t)
\end{equation}
\begin{equation}
\label{train_2}
\frac{\partial \log P(x,z)}{\partial \Theta_l} = \frac{\partial}{\partial \Theta_l} \sum_{t=1}^T \log P(z_t|\mathcal A_t)
\end{equation}
where $\Theta_a,\Theta_l$ are the parameters of the acoustic and language models respectively. 

\section{Inference}

In the hybrid architecture, the prediction $z_t$ at time $t$ is conditioned upon the entire sequence history $\mathcal A_t$ due to the RNN language model. This enforces successive frames to be coherent and thus performs temporal smoothing. In addition to temporal smoothing, an accurate language model can impose musicological rules and restrictions on the output transcriptions. While decoding, proceeding in a greedy chronological manner yields sub-optimal results because the sequence history $\mathcal A_t$ has not been optimally determined. At the same time, exhaustively searching for the globally optimal sequence is intractable since each non-leaf node in the search graph has $2^{N}$ descendants. Instead, we perform a global search for the most likely sequence using beam search, a breadth-first tree search algorithm that keeps track of only the $w$ most promising paths at any depth $t$ \cite{graves2012sequence,boulanger2013high,boulangerphone}. In the search graph, a node at depth $t$ corresponds to a subsequence of 
length $t$ and the log-likelihood of each sub-sequence is the heuristic that guides search. 

In addition to the beam width $w$, the high-dimensional variant of the beam-search algorithm outlined in \cite{boulanger2013high} requires an additional parameter, the branching factor $K$. When using complex distribution estimators like the NADE, deterministically enumerating all possible configurations in order of decreasing probability is intractable. In such situations, the algorithm proceeds by making a pool of the top $K$ candidate solutions by sampling. Random sampling from the conditional distribution of the language model is slow and inefficient and limits the size of the beam width during search. 




\begin{algorithm}
\caption{High Dimensional Beam Search \cite{boulanger2013high}}
\begin{algorithmic}
\State{Find the most likely sequence $z$ given $x$ with a beam width $w$.}
\State{$q \gets$ min-priority queue}
\State{$q$.insert$(0,\left\{\right\},m_{lm},m_{am})$}
\For{$t$ = $1$ to $T$}
	\State{$q' \gets$ min-priority queue of capacity $w$ $\ast$}
	\While{$q'$.len() $< w$}
		\For{$l,s,m_{lm},m_{am}$ \textbf{in} $q$}
			\State{$z' = m_{am}$.next\_most\_probable()}
      \State{$l' = \log P_{lm}(z'|s)P_{am}(z'|x)-\log P(z')$}
			\State{$m_{lm}' \gets m_{lm}$ with $z_t := z'$}
			\State{$m_{am}' \gets m_{am}$ with $x := x_{t+1}$}
			\State{$q$.insert($l+l',\left\{ s,z' \right\},m_{lm}',m_{am}')$}
		\EndFor
	\EndWhile
	\State{$q \gets q'$}
\EndFor
\Return{$q.$pop()}
\State{$\ast$ A \emph{min}-priority queue of capacity $w$ maintains the $w$ highest values at all times.}
\end{algorithmic}
\end{algorithm}

Instead of pooling the top $K$ configurations by drawing samples from the language model at each time step, we propose using the acoustic model to enumerate the most likely predictions. The motivation for doing so is twofold. Firstly, using the most likely solutions from the acoustic model to direct search avoids cases where the language model makes mistakes early on in a sequence and can never recover from them. Secondly, the outputs of the acoustic classifier are independent of each other. Enumerating the most likely solutions with a DP algorithm is more efficient than stochastic sampling \cite{boulanger2013high}. Unlike \cite{boulanger2013high}, the high-dimensional beam search algorithm outlined in algorithm 1 does not require the branching factor $K$ to be specified in advance and allows the use of much larger beam widths.

\section{Experiments}
\subsection{Acoustic Modelling}

We experiment with using 3 different neural network architectures for learning relevant features from spectrogram inputs. Firstly, we use a deep, feed-forward neural network (DNN) as the acoustic classifier. DNNs currently form the state of the art for acoustic modelling in speech \cite{hinton2012deep} and have been successfully applied to music transcription in the past \cite{nam2011classification,boulanger2012modeling}. The ability of DNNs to learn a hierarchy of increasingly complex features makes them an ideal choice for acoustic modelling. 

Despite being powerful frame-level classifiers, DNN outputs are often noisy because they do not account for dependencies between input frames. In order to avoid this issue, we also experiment with using an RNN acoustic model. DNNs base their predictions upon a single frame of input, while the predictions of an RNN at time $t$ are conditioned on all frames for time $\tau < t$. Previous work on using RNNs as acoustic models for transcription demonstrates that RNNs are very good at predicting note-onsets \cite{bock2012polyphonic}. We use the stacked RNN architecture, where several recurrent hidden layers are stacked in order to encourage each recurrent layer to operate at a different timescale \cite{schmidhuber1992learning}. One limitation of using the RNN as the acoustic model is that it violates the independence assumption made in Equation \ref{independence}. The RNN predictions at $t$ are conditioned on all past inputs for $\tau < t$ through the hidden layers. Since the language model and the acoustic model 
are trained separately, combining their predictions leads to certain factors being counted twice. Although in theory, this makes it hard to use RNN acoustic models, in our experiments we discovered that this difficulty does not affect performance. 

Finally, we experiment with using the features learnt by a DNN as inputs to an RNN. The motivation for doing this is that the features learnt by the DNN are believed to disentangle the factors of variation present in the inputs \cite{goodfellow2009measuring}. It is easier for the RNN to discover relationships between frames of disentangled features as compared to the original spectrogram inputs. We use the activations of the hidden units of all the layers of a DNN as input features to a stacked RNN. 
\subsection{Language Modelling}

As mentioned in Section 2, the RNN can be used as a generative model to define distributions over sequences. Unlike speech recognition, where the language model computes a multinomial distribution over a discrete set of phoneme labels, the MLM has to compute distributions over high-dimensional binary vectors. In order to capture the interactions between the output variables at each time-step, we prefer to use the RNN-NADE over the RNN as the MLM. At each step, the conditional NADE defines a joint distribution over the space of high-dimensional binary output vectors. At test time, the conditional NADE at time $t$ provides the likelihood of observing the vectors predicted by the acoustic model, conditioned on all the predictions so far.  
\subsection{Experimental Setup}

\begin{table*}[htpb]
\begin{center}
  \begin{tabular}{|c| c  c | c  c | c  c | c c|}
    \hline
    Post Processing& \multicolumn{2}{|c|}{None}&\multicolumn{2}{c|}{Thresholding}&\multicolumn{2}{c|}{HMM}&\multicolumn{2}{c|}{Hybrid Architecture}\\ \hline
    Acoustic Model& Frame & Note & Frame & Note & Frame & Note & Frame & Note \\ \hline
    DNN & 66.33 & 56.09 & 67.95 &59.58 &68.16 & 62.50 & 69.25 & 62.90 \\ \hline
    RNN & 66.83 & 61.48 & 67.92 & 62.40 & 67.27 & 65.36 &68.24 & \textbf{67.4} \\ \hline
    DNN + RNN & 68.83 &62.41 &69.30 &61.35 & 68.60 & 63.45 & \textbf{69.62} &64.69 \\ \hline
  \end{tabular}
\end{center}
\vspace{-1.8em}\caption{F-measures for multiple pitch detection on the MAPS dataset}
\end{table*}
We perform experiments on the MAPS dataset \cite{emiya2010multipitch} to test the performance of the hybrid architecture and compare its performance to other models. The MAPS dataset consists of 270 pieces of piano music along with their ground truth MIDI transcriptions. 210 of these are rendered by software synthesisers, while 60 are played on real pianos. 
\begin{table}[htbp]
\scalebox{0.9}{
\begin{tabular}{|c|c c|c c|c c|}
 \hline 
&\multicolumn{2}{|c|}{Precision} &\multicolumn{2}{|c|}{Recall}&\multicolumn{2}{|c|}{Accuracy} \\ \hline
Acoustic Model&Frame&Note&Frame&Note&Frame&Note \\ \hline
DNN & \textbf{66.61} &61.37 &72.12&64.52&52.97&45.88 \\ \hline
RNN & 62.41 &\textbf{66.25} &75.28&\textbf{68.6}&51.79&\textbf{50.83} \\ \hline
DNN+RNN & 63.18 &65.57 &\textbf{77.51}&63.84&\textbf{53.39}&47.81 \\ \hline
\end{tabular}
}
\caption{Additional evaluation metrics for the hybrid architecture.}
\end{table}
For our experiments, we randomly select 200 tracks for training, 20 for validation and 50 for testing\footnote{Training/testing data info at: www.eecs.qmul.ac.uk/\texttildelow sss31/}. We use the entire length of the training and validation tracks and use the first 30 seconds of the tracks for testing. Pre-processing the data consisted of downsampling the tracks to $16$ kHz and calculating the magnitude spectrogram. Spectrograms were computed with a window size of $64$ ms and a hop size of $32$ ms for the training and validation tracks. For the test tracks, spectrograms were computed every $10$ ms \cite{bay2009evaluation}. The spectrograms were further preprocessed by subtracting the mean and dividing by the standard deviation of each frequency bin, calculated over the training set.

\subsection{Training}

The acoustic and language models were trained by gradient descent, according to Equations \ref{train_1} and \ref{train_2}. The output layers of both the DNN and RNN acoustic models consisted of sigmoid units. Each output of the acoustic model can be interpreted as the independent probability of a pitch being present in that frame. The acoustic classifiers were trained by minimising a cross entropy cost, since the target vectors for all frames are high-dimensional binary vectors. For both DNN and RNN models, weights were randomly initialised by sampling values from a Gaussian distribution with $0$ mean and $0.01$ standard deviation. We also used a momentum of $0.9$ while updating the weights. The DNN models were trained on independent frames of spectrograms extracted from the training set. For training the stacked RNN models, the training tracks were further divided into sub-sequences of length 200 and the models were trained by Back-Propagation Through Time (BPTT) \cite{RumelhartHintonWIlliams1986}. The RNN-NADE language models were trained on the ground truth MIDI data associated with the training data. The RNN-NADE models were optimised with Hessian Free (HF) optimisation. 

\subsection{Evaluation Metrics}

We evaluate the performance of our system using the evaluation metrics used in MIREX \cite{bay2009evaluation}. We present F-measures for both frame-based and onset-only note-based tracking evaluation metrics. Additionally, we report precision, recall and accuracy measures for the 3 best performing models.

\section{Results}
In Table 1, we present F-measures for the different systems evaluated using different combinations of acoustic models and post-processing. We report F-measures for both frame-based and note onset based evaluation metrics \cite{bay2009evaluation}. The best DNN acoustic model consists of $3$ layers with $100$ units each. The RNN acoustic models have two stacked hidden layers with $250$ hidden units each. For language modelling, the conditional NADEs have $150$ hidden units and the RNN has $100$ hidden units. Four types of post-processing are considered in the experiments. No post processing, where the most likely outputs from the classifiers are chosen; learning independent thresholds for each classifier output based on the training set; HMM post processing assuming each pitch-class is independent; and finally the proposed hybrid architecture with a beam width $w = 100$. The post processing also includes minimum duration pruning ($70$ ms) to improve the model's accuracy at detecting note-onsets. 

From Table 1, we observe that the hybrid architecture consistently outperforms other methods. The best F-measure on both frame-based and note-onset based metrics is achieved by the hybrid architecture. The note-onset based F-measure is comparable to the frame-based F-measure which demonstrates the ability of the model to accurately identify note onsets. Beam search post-processing leads to a $3\%$ increase in frame-based F-measure and a $6\%$ increase in note-onset F-measure over greedy search ($w = 1$) for the DNN acoustic model. The RNN acoustic models are better at accurately predicting note-onsets because they implicitly perform temporal smoothing. In our experiments we discovered that the noisy DNN outputs when smoothed with a median filter, performed equally well as the RNN acoustic models on the note-based metrics. The relative improvement in performance when using the hybrid architecture is maximum for the DNN acoustic models, which is probably due to the fact that they do not violate the independence assumption in Equation \ref{independence}.  Table 2 shows additional metrics for the 3 hybrid models that perform best. It is clear that most of the errors are due to false alarms, which can be attributed to the error in accurately modelling note durations. However this error is not unique to this particular system and persists even in the ground truth transcriptions. The beam search takes 20 hours on a CPU to decode the first 30 seconds of all the test tracks. 

\section{Conclusion}

We present a hybrid RNN-based architecture for including symbolic priors in an automatic music transcription system. The architecture combines acoustic and high-level symbolic predictions in a principled manner and we propose an efficient algorithm for inference. The model generalises the popular technique of using independent HMMs to smooth the predictions of acoustic classifiers. Evaluation on the MAPS dataset suggests that the model outperforms related music transcription systems. In the future, we plan to work on improving the individual components of the architecture, namely the acoustic and language modeling. We would also like to investigate ways to improve beam search to make it feasible for real-time applications. Finally, we would like to expand our evaluations to datasets with multiple instruments. 


\bibliography{biblio}{}
\bibliographystyle{plain}
\end{document}